\documentclass[10pt,journal,compsoc]{IEEEtran}
\usepackage{times}
\usepackage{latexsym}
\usepackage{pifont}
\usepackage[normalem]{ulem}
\usepackage{gensymb}
\usepackage{footmisc}

\usepackage{graphicx}
\usepackage{epsfig}
\usepackage{epstopdf}
\usepackage{rotating}
\usepackage{wrapfig}
\usepackage{float}

\usepackage{amsmath}
\usepackage{amssymb}

\usepackage{framed}
\usepackage{booktabs}
\usepackage{multirow}
\usepackage{multicol}
\usepackage{array}
\usepackage{longtable}
\usepackage{tabularx}
\usepackage{collcell}
\usepackage{xcolor}
\usepackage{colortbl}
\PassOptionsToPackage{table}{xcolor}

\usepackage{enumitem}

\usepackage{url}

\usepackage{backnaur} 
\usepackage{hhline}
\usepackage{listings}
\usepackage{fancyvrb}

\useunder{\uline}{\ul}{}

\definecolor{Gray}{gray}{0.94} 

\newcommand*{\MinNumber}{-0.01}%

\newcommand{\AG}[1]{%

        \ifdim #1 pt > \MinNumber pt
            \textcolor{black}{#1}

        \fi

}

\ifCLASSOPTIONcompsoc
  \usepackage[nocompress]{cite}
\else
  \usepackage{cite}
\fi

\ifCLASSINFOpdf
\else
\fi

\newcommand{\caim}{CAIM{} }
\newcommand{\hfr}{\textit{HFR}{} }

\newcommand{\fullcaim}{Conditional Adaptive Instance Modulation{} }
\newcommand{\etal}{\textit{et al.}}

\begin{document}

\title{From Modalities to Styles: Rethinking the Domain Gap in Heterogeneous Face Recognition}
\author{Anjith~George,~\IEEEmembership{Member,~IEEE,}
        S\'ebastien~Marcel,~\IEEEmembership{Senior~Member,~IEEE}
\IEEEcompsocitemizethanks{\IEEEcompsocthanksitem All authors are with Idiap Research Institute, Martigny, Switzerland. S\'{e}bastien Marcel is also affiliated with Universit\'{e} de Lausanne (UNIL), Lausanne, Switzerland.
E-mail: \{anjith.george, sebastien.marcel\}@idiap.ch 
}
\thanks{Manuscript received April 19, 2005; revised August 26, 2015.}}

\markboth{Journal of \LaTeX\ Class Files,~Vol.~14, No.~8, August~2015}%
{Shell \MakeLowercase{\textit{et al.}}: Bare Demo of IEEEtran.cls for Biometrics Council Journals}

\IEEEtitleabstractindextext{%
\begin{abstract}

Heterogeneous Face Recognition (HFR) focuses on matching faces from different domains, for instance, thermal to visible images, making Face Recognition (FR) systems more versatile for challenging scenarios. However, the domain gap between these domains and the limited large-scale datasets in the target HFR modalities make it challenging to develop robust HFR models from scratch. In our work, we view different modalities as distinct styles and propose a method to modulate feature maps of the target modality to address the domain gap. We present a new \fullcaim (\caim\!) module that seamlessly fits into existing FR networks, turning them into HFR-ready systems. The \caim block modulates intermediate feature maps, efficiently adapting to the style of the source modality and bridging the domain gap. Our method enables end-to-end training using a small set of paired samples. We extensively evaluate the proposed approach on various challenging HFR benchmarks, showing that it outperforms state-of-the-art methods. The source code and protocols for reproducing the findings will be made publicly available.

\end{abstract}

}

\maketitle

\IEEEdisplaynontitleabstractindextext

%
\IEEEpeerreviewmaketitle

\IEEEraisesectionheading{\section{Introduction}\label{sec:introduction}}

\IEEEPARstart{F}acial recognition (FR) technology has gained popularity in the field of access control due to its high efficiency and user-friendly nature. Most state-of-the-art FR methods achieve excellent performance in ``in the wild'' conditions and even reach a level comparable to human performance in recognizing faces \cite{learned2016labeled}, thanks to the advancement of convolutional neural networks (CNN).
Typically, FR systems are designed to work within a homogeneous domain, meaning that both the enrollment and matching phases are conducted using the same type of data, usually facial images captured with an RGB camera. Nonetheless, there are scenarios where performing matching in a heterogeneous setting could be beneficial. For instance, near-infrared (NIR) cameras, commonly found in smartphones and security cameras, offer superior performance across various lighting conditions and exhibit resilience to spoofing attacks \cite{li2007illumination,george2022comprehensive}. Despite these advantages, developing an FR system tailored for NIR imagery requires an extensive collection of annotated training data, which is often scarce.
\begin{figure}[t!]
    \centering
    \includegraphics[width=0.85\linewidth]{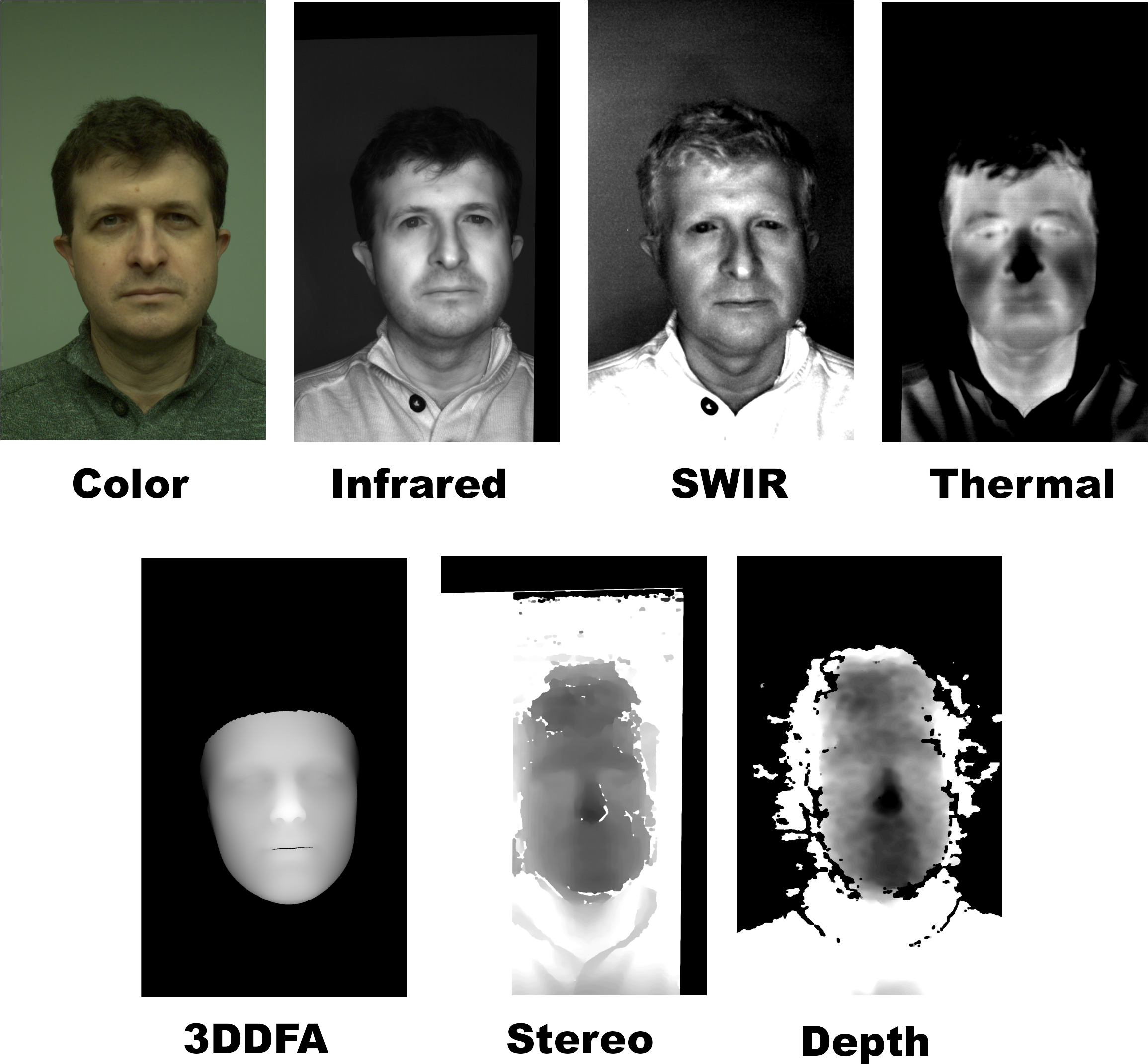}
    \caption{This figure shows the facial images of the same individual acquired using distinct imaging modalities (Images taken from  MCXFace dataset \cite{george2022prepended}). The task in \hfr is to facilitate cross-domain matching while overcoming the challenges posed by the domain gap.} \label{fig:hfr}
  \end{figure}

Heterogeneous Face Recognition (\hfr\!) systems are designed to facilitate cross-domain matching (Fig. \ref{fig:hfr}), enabling the comparison of enrolled RGB images with NIR (or other types of) images without necessitating the enrollment of separate modalities \cite{klare2012heterogeneous}. This approach proves to be invaluable, especially in conditions where acquiring visible images is challenging. For example, thermal images can be utilized for recognition purposes regardless of the lighting conditions, making face recognition feasible day and night, and even at considerable distances. \hfr systems are versatile and capable of processing and matching facial images from diverse sources and modalities, which significantly broadens the potential applications and utility of face recognition systems across various challenging scenarios.

\hfr extends the use of Face Recognition (FR) systems to challenging scenarios, such as those involving low-lighting or long-range, by capitalizing on the specific characteristics of imaging modalities. HFR approaches effectively mitigate certain constraints, broadening the scope and applicability of FR systems. Despite its usefulness, developing a  Heterogeneous Face Recognition (\hfr\!) system comes with its own challenges. Cross-domain matching is challenging primarily because of the domain gap. This gap can lead to a drop in performance when face recognition (FR) networks, which are typically trained on visible-light images, are applied to images from different sensing modalities \cite{he2018wasserstein}. Moreover, creating models that are robust to both visible and other modalities is challenging, exacerbated by the limited availability of large-scale multimodal datasets. Collecting large-scale paired datasets for these additional modalities is not only challenging but can also incur significant costs. Hence, it is essential to devise an HFR framework that requires only a limited set of labeled samples for training the models.

In our approach, we build upon face recognition networks pre-trained with a large dataset of faces from the visible spectrum, using it as our foundational network. We address the challenge of different modalities by conceptualizing them as unique \textit{styles}. Our proposed framework is designed to bridge the domain gap by adapting the network's intermediate feature maps to align with these styles. The core of our method is the introduction of a new module, which we refer to as the  \fullcaim (\caim\!) \cite{george2023bridging}. The CAIM module can be integrated seamlessly into the face recognition network's intermediate stages. This trainable module is capable of being trained from scratch, transforming a standard face recognition system into an HFR network capable of handling a variety of modalities, all while requiring a minimal number of training sample pairs.

The main contributions of this work are as follows:

  \begin{itemize}

    \item We conceptualize the domain gap in Heterogeneous Face Recognition (HFR) as a manifestation of distinct \textit{styles} from different imaging modalities, and address this domain gap as a style modulation problem. 
          
    \item A new trainable component called \fullcaim (\caim\!) is introduced, which can transform a pre-trained FR network into a heterogeneous face recognition network, requiring only a limited number of paired samples for training. 
    
    \item We implemented our approach with two different face recognition models to evaluate the generalization of our approach.
    
    \item We demonstrate the robustness and effectiveness of our proposed method through extensive evaluation on various challenging HFR benchmarks.

  \end{itemize}

  Finally, the protocols and source codes will be made available publicly \footnote{\url{https://gitlab.idiap.ch/bob/bob.paper.ijcb2023_caim_hfr}}.

The structure of the rest of the paper is organized in the following manner: In Section \ref{sec:related_work}, we review the previous literature in Heterogeneous Face Recognition (HFR). The specifics of the CAIM approach are elaborated in Section \ref{sec:approach}. Section \ref{sec:experiments} and \ref{sec:discussions} provide a thorough evaluation of the CAIM method, including comparative analyses with state-of-the-art methods, followed by in-depth discussions. Finally, Section \ref{sec:conclusions} concludes the paper with a summary of our findings and proposes directions for future research.

\begin{figure*}[t!]
  \centering
  \includegraphics[width=0.90\linewidth]{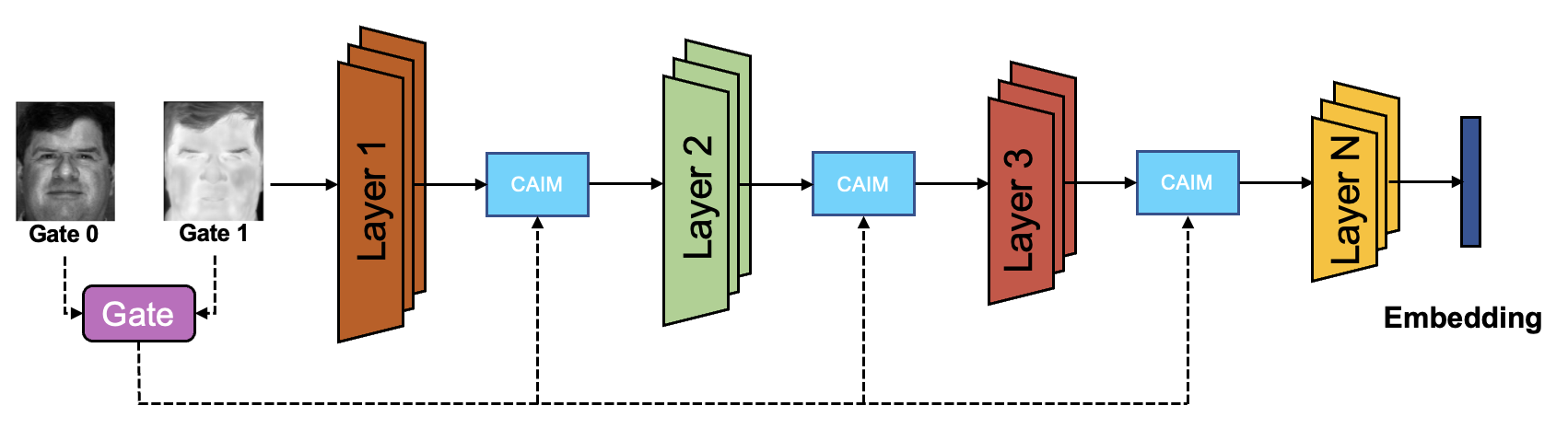}
  \caption{Schematic diagram of the proposed framework: Layer 1 to Layer N represent the frozen blocks of layers from a pretrained Face Recognition (FR) model. The \caim module is inserted between the initial few blocks. }
  \label{fig:framework}
\end{figure*}

\section{Related work} 
\label{sec:related_work}

The objective of Heterogeneous Face Recognition (HFR) methods is to accurately match faces across images captured by different sensing modalities. Yet, the discrepancy between these domains, known as the domain gap, can impair the efficacy of face recognition networks when performing a direct comparison of multi-modal images. Therefore, it's crucial for HFR methodologies to effectively close this modality gap. In this section, we review recent literature on strategies proposed for addressing the domain gap.

\subsection{Invariant feature-based methods}

Various strategies have been developed for Heterogeneous Face Recognition (HFR) with the goal of extracting features that remain consistent across different imaging modalities. Liao \etal \cite{liao2009heterogeneous} introduced a technique that relies on the Difference of Gaussian (DoG) filters combined with multi-scale block Local Binary Patterns (MB-LBP) to capture invariant features. Klare \etal \cite{klare2010matching} proposed a method employing Local Feature-based Discriminant Analysis (LFDA), which utilizes Scale-Invariant Feature Transform (SIFT) and Multi-Scale Local Binary Pattern (MLBP) as feature descriptors. Zhang \etal \cite{zhang2011coupled} proposed the Coupled Information-Theoretic Encoding (CITE) approach, which seeks to maximize the mutual information across modalities within quantized feature spaces. Approaches based on Convolutional Neural Networks (CNNs) have also been applied to HFR, demonstrating the versatility of deep learning models in this context \cite{he2017learning,he2018wasserstein}. Roy \etal \cite{roy2018novel} proposed a method termed Local Maximum Quotient (LMQ), specifically designed to identify invariant features in cross-modal facial imagery. In \cite{liu2018composite}, authors introduced a feature-based approach for HFR for composite sketch recognition. This approach involved extracting features using the Scale-Invariant Feature Transform (SIFT) and the Histogram of Oriented Gradient (HOG) from different facial components. These features were then integrated at the score level, where the facial components were combined using a linear function.

\subsection{Common-space projection methods}

Common-space projection methods aim to learn a transformation that projects facial images from various modalities into a unified subspace, thereby reducing the domain gap \cite{kan2015multi,he2017learning}. Lin and Tang \cite{lin2006inter} devised a method known as common discriminant feature extraction to extract features from cross-modal images and align them within a shared feature space. Yi \etal \cite{yi2007face} utilized Canonical Correlation Analysis (CCA) to correlate Near-Infrared (NIR) and Visible Spectrum (VIS) face images. Lei \textit{et al.} \cite{lei2009coupled,lei2012coupled} introduced regression-based techniques to establish mapping functions that bridge the gap between different modalities. Sharma and Jacobs \cite{sharma2011bypassing} developed a method based on Partial Least Squares (PLS) to learn a linear mapping that maximizes the covariance between face images across modalities. Klare and Jain \cite{klare2012heterogeneous} proposed a method for representing face images by their similarities to a predefined set of prototype faces, followed by projecting these representations onto a linear discriminant analysis subspace for recognition purposes. In \cite{de2018heterogeneous}, authors suggested that the high-level features in convolutional neural networks, when trained on visible light spectra, are actually domain-agnostic and can be used to encode images from other sensing modalities. They adapt the initial layers of a pre-trained FR model, termed Domain-Specific Units (DSUs), to minimize the domain gap, training the entire system in a contrastive learning setting. A challenge with this approach is determining the exact number of layers to adapt, which requires thorough experimentation to optimize. Liu \textit{et al}.  \cite{liu2020coupled} proposed a novel method known as Coupled Attribute Learning for HFR (CAL-HFR), which uniquely does not require manual labeling of facial attributes. This method utilizes deep convolutional networks to map face images from heterogeneous scenarios into a shared space. Additionally, they introduced the Coupled Attribute Guided Triplet Loss (CAGTL), a specially designed loss function aimed at addressing the issues of inaccurately estimated attributes in end-to-end training. Recently, Liu \textit{et al}. \cite{liu2023modality} proposed a semi-supervised learning approach for modality-independent heterogeneous face recognition (HFR) representation, termed as Modality-Agnostic Augmented Multi-Collaboration representation for Heterogeneous Face Recognition (MAMCO-HFR). This method introduces a multi-collaborative face representation that leverages interactions across various network depths to extract potent discriminative information for identity recognition. Additionally, they proposed a modality-agnostic augmentation technique that creates adversarial disturbances to effectively map unlabeled faces into a modality-agnostic domain.

\subsection{Synthesis based methods}

Synthesis-based approaches in Heterogeneous Face Recognition (HFR) \cite{tang2003face,fu2021dvg} focus on creating images in the source domain from those in the target modality. This synthetic generation facilitates the use of standard face recognition networks for biometric identification. Wang \etal \cite{wang2008face} explored a patch-based synthetic method utilizing Multi-scale Markov Random Fields, and Liu \etal \cite{liu2005nonlinear} employed Locally Linear Embedding (LLE) for establishing a pixel-wise correspondence between visible (VIS) images and viewed sketches. The use of CycleGAN, as presented in  \cite{zhuUnpairedImagetoImageTranslation2017}, for unpaired image-to-image translation has paved the way for transforming target domain images to match the source domain \cite{baeNonvisualVisualTranslation2020}. Furthermore, Zhang \etal \cite{zhang2017generative} introduced a method using Generative Adversarial Networks (GANs) to create photo-realistic VIS images from polarimetric thermal images through GAN-based Visible Face Synthesis (GAN-VFS). Several recent approaches have been proposed using GANs for the synthesis of VIS images from another modality, such as the Dual Variational Generation (DVG-Face) framework \cite{fu2021dvg}, which achieved state-of-the-art results in many challenging \hfr benchmarks. Liu \textit{et al}. \cite{liu2021heterogeneous} introduced the Heterogeneous Face Interpretable Disentangled Representation (HFIDR), a novel approach capable of explicitly interpreting the dimensions of face representation. This method focuses on extracting latent identity information for cross-modality recognition and employs a technique to transform the modality factor, enabling the synthesis of cross-modality faces. In \cite{luo2022memory}, authors proposed the Memory-Modulated Transformer Network (MMTN) for HFR, treating the problem as an unsupervised, reference-based ``one-to-many'' generation problem. The MMTN incorporates a memory module to capture prototypical style patterns and a style transformer module to blend the styles of input and reference images at a local level. Recently, George \etal \cite{george2022prepended} introduced the concept of Prepended Domain Transformers (PDT), which prepends a trainable neural network module to a pre-trained FR network, converting it into an HFR network. This module translates feature representations to align cross-domain embeddings in the feature space, without the need for explicit generation of source domain images.

\subsection{Challenges in HFR}

In recent literature, Heterogeneous Face Recognition (HFR) methods, particularly those utilizing Generative Adversarial Networks (GANs), have gained prominence for their synthesis-based approaches. These methods, such as DVG-Face and GAN-VFS \cite{fu2021dvg,zhang2017generative} achieve reasonable results in generating high-fidelity images. Leveraging a pre-trained FR model in such synthesis-based HFR methods obviates the requirement for a vast amount of training data to develop a new FR model. Nonetheless, the synthesis step introduces a significant computational burden, which may hinder its practical deployment in real-life scenarios. We propose a different perspective: treating the domain gap between visible images and images from other modalities as a variation in ``styles''. By adopting this viewpoint, we can address the domain gap directly within the feature space through modulation of the feature maps. This strategy eliminates the computational and memory-intensive process of synthesizing images in the source modality.

\section{Proposed Method}
\label{sec:approach}

We follow the notations consistent with recent literature \cite{weiss2016survey,de2018heterogeneous, george2022prepended, george2024diu} to formally define the \hfr task.

\subsection{Formal definition of \hfr}

Consider a domain $\mathcal{D}$ that includes a set of samples $X \in \mathbb{R}^d$ and a marginal distribution $P(X)$ (of dimensionality $d$). The goal of a face recognition (FR) system, $\mathcal{T}^{fr}$, can be characterized by a label space $Y$ with a conditional probability $P(Y|X,\Theta)$, where $X$ and $Y$ represent random variables, and $\Theta$ denotes the parameters of the model. In the training stage of an FR system, the conditional probability $P(Y|X, \Theta)$ is typically determined through supervised learning using a face dataset $X={x_1, x_2, ..., x_n}$ and their corresponding identity labels $Y={y_1, y_2, ..., y_n}$.

In the heterogeneous face recognition (\hfr\!) problem, we assume the presence of two domains: a source domain $\mathcal{D}^s = {X^s, P(X^s)}$ and a target domain $\mathcal{D}^t = {X^t, P(X^t)}$, both sharing the labels $Y$. The objective of the \hfr problem,  $\mathcal{T}^{hfr}$, is to estimate a  $\hat{\Theta}$ such that $P(Y|X^s, \Theta) = P(Y|X^t, \hat{\Theta})$.

\subsection{Proposed Framework}

In our proposed approach, we consider face images from various modalities as separate \textit{styles}, considering the domain discrepancy in the \hfr challenge to be a manifestation of these style variations. We propose that by addressing the domain-specific \textit{style}, we can reduce the domain gap. To accomplish this, we employ conditional modulation on the intermediate feature maps within a pre-trained face recognition network.

Using the parameters $\Theta_{FR}$ from a pre-trained face recognition (FR) model developed on the source domain $\mathcal{D}^s$, our strategy does not alter the model's original weights. Instead, we inject a set of trainable network modules between the frozen layers of the FR network, named \caim\!, which are designed to modulate the intermediate feature maps. The \caim modules perform normalization and style modulation on feature maps from the target modality, to align the embeddings of corresponding samples from both modalities in the embedding space. Figure \ref{fig:framework} illustrates the overall design of our proposed system. We incorporate the \caim modules primarily within the initial blocks of the network, as these are more closely related to modality-specific characteristics. An external gating mechanism is deployed to enable the \caim modules solely for the target modality data while allowing the source modality data to pass unaffectedly, thereby mitigating the risk of catastrophic forgetting.

The \hfr task can be mathematically formulated as:
\begin{equation}
  P(Y|X_t, \hat{\Theta}) = P(Y|X_t, [\Theta_{FR}, \theta_{\caim_{i,  i \in (1,2,..,K)}}])
\end{equation}

Where, $\theta_{\caim_{i}}$ denotes the $i^{th}$ \caim block out of $K$ blocks.

The \caim blocks, specified by the parameters $\theta_{\caim_{i}}$, are the only trainable components and can be fine-tuned in a supervised manner. When the system is trained, the \caim module acts as a pass-through for images from the source domain ($X^s$), effectively allowing the network to produce the reference embeddings through $\Theta_{FR}$. However, for images from the target domain ($X^t$), the processing involves both the frozen network layers ($\Theta_{FR}$) and the newly introduced \caim blocks. The training utilizes a contrastive loss function, as described by \cite{hadsell2006dimensionality}, to align the embeddings in the shared representational space. The contrastive loss is given as:

\begin{equation}
\begin{split}
\mathcal{L}_{Contrastive}(\hat{\Theta}, Y_p, X_s, X_t)= & (1-Y_p)\frac{1}{2}D_W^{2}  \\
            & + Y_p\frac{1}{2}{max(0, m-D_W)}^{2}
\end{split}
\end{equation}

Where $\hat{\Theta}$ represents the network's weights together with the frozen weights, $X_s$ and $X_t$ denote heterogeneous image pairs. The label $Y_p$ indicates whether the pairs share the same identity. The margin in the contrastive loss function is denoted by $m$, while $D_W$ represents the metric used to compute the distance between the embeddings of the two images in a pair. The chosen distance measure $D_W$ could be the Euclidean distance or the cosine distance, depending on which is used to compare the feature representations produced by the network. Further details on the \caim block's design are provided in subsequent subsections.

\subsection{Architecture of the \caim block}

\begin{figure*}[t!]
  \centering
  \includegraphics[width=0.79\linewidth]{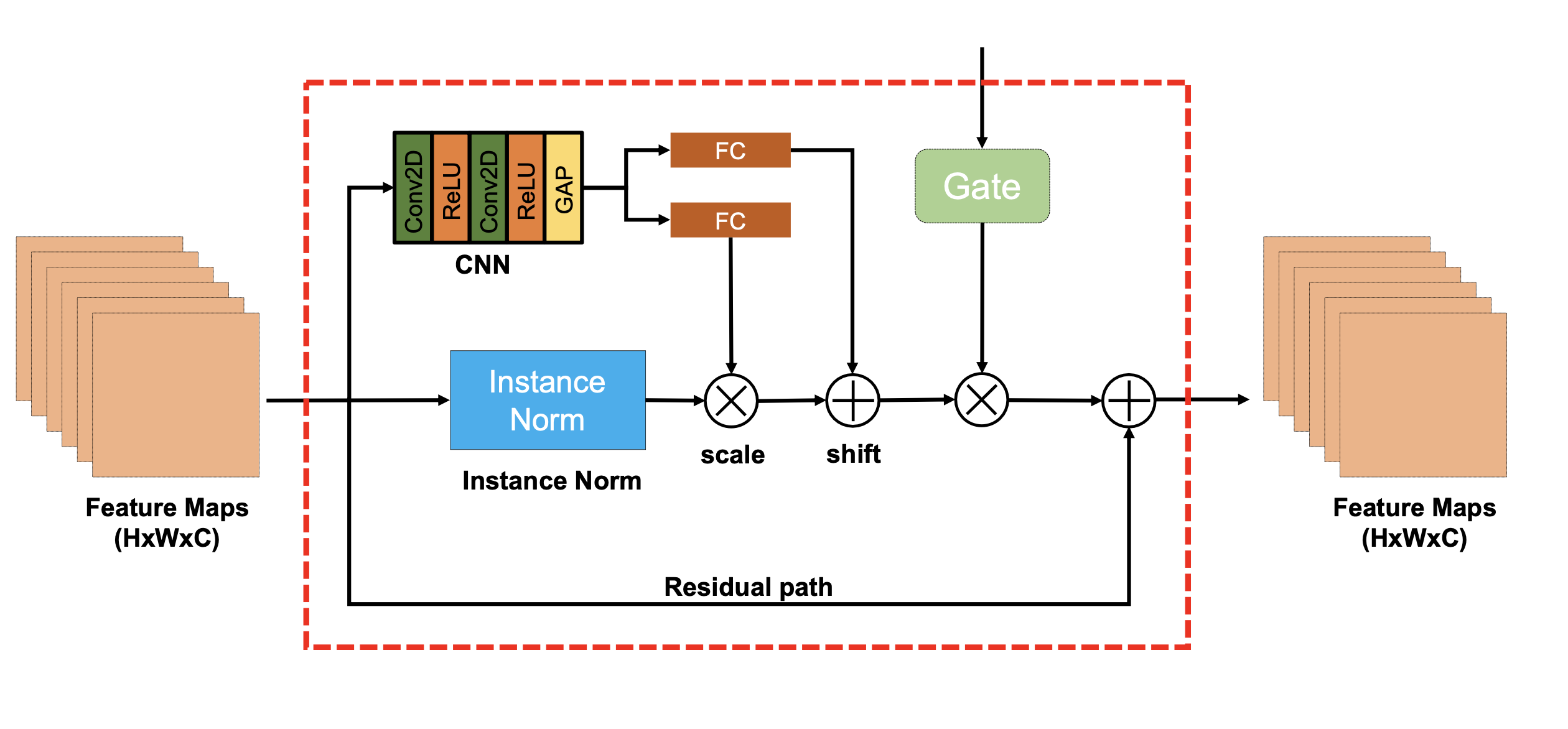}
  \caption{Architecture of the \fullcaim (\caim\!) block. The global gate signal activates the block. The gate signal becoming zero deactivates this module and the entire module functions as an identity block in this case due to the residual path.}
  \label{fig:caim_arch}
\end{figure*}

Figure \ref{fig:framework} illustrates the components of the proposed \fullcaim framework. The \caim blocks are inserted between the frozen layers of the pre-trained face recognition network. The detailed design of the \caim block is shown in Fig. \ref{fig:caim_arch}. This block takes an input feature map along with a global gating signal and outputs a feature map with the same dimensions. The first component in the \caim block is an Instance Normalization (IN) layer that normalizes each feature map individually, without trainable affine parameters. Following this, a parallel Convolutional Neural Network (CNN) module takes the original, un-normalized feature map to extract a shared representation. This CNN module consists of two sets of $3\times3$ convolutional layers, each followed by a Rectified Linear Unit (ReLU) activation function, and then a Global Average Pooling (GAP) layer. To generate the scaling and shifting parameters for the normalized feature maps, two dense (fully connected) layers are appended to the shared representation. These dense layers are tasked with calculating the parameters that modulate the normalized feature maps. Furthermore, a residual connection is incorporated into the network. When the global gate signal is set to zero, the \caim block acts as an identity function, obtaining the same embeddings from the original pre-trained face recognition network for the reference modality, effectively bypassing the modulation process.

\subsection{Style Modulation for HFR}

In this section, we discuss using style modulation as a strategy to bridge the domain gap between visible and other modalities, starting with the application of Instance Normalization \cite{ulyanov2017improved}.

\subsubsection{Instance normalization}

Previous works have demonstrated that the statistical properties of feature maps in deep neural networks (DNNs) effectively capture the style of images \cite{gatys2016image}. Ulyanov \etal \cite{ulyanov2017improved} showed that substituting batch normalization layers with Instance Normalization (IN) significantly enhances style transfer. Instance Normalization layer normalizes the feature maps, and this process can be represented as follows:

\begin{equation}
\mathrm{IN}(x) = \gamma \left(\frac{x - \mu(x)}{\sigma(x)}\right) + \beta
\end{equation}

Here, $\gamma, \beta \in \mathbb{R}^{C}$ represent affine parameters learned from the data, while $\mu(x)$ and $\sigma(x)$ are calculated across spatial dimensions for each individual sample, as opposed to across mini-batches in BatchNorm.

Dumoulin \etal \cite{dumoulin2016learned} subsequently extended Instance Normalization to Conditional Instance Normalization (CIN). In this approach, a set of parameters, $\gamma^{s}$ and $\beta^{s}$, can be learned for a predefined set of styles, denoted by $s$.

In \cite{huang2017arbitrary}, Huang \etal introduced a network module named Adaptive Instance Normalization (AdaIN), specifically designed for image style transfer. This module is designed to align the mean and variance of content features with those of style features in the context of image style transfer. The authors suggest that Instance Normalization facilitates style normalization by normalizing the feature statistics, namely the mean and variance. The AdaIN module operates by taking a content input $x$ and a style input $y$ and aligning the channel-wise mean and variance of $x$ to correspond with those of $y$. Unlike other methods, AdaIN does not utilize learnable affine parameters; rather, it dynamically computes these parameters based on the style input.

\begin{equation}
\mathrm{AdaIN}(x, y) = \sigma(y) \left(\frac{x - \mu(x)}{\sigma(x)}\right) + \mu(y)
\end{equation}

The normalized content input is scaled by $\sigma(y)$ and shifted by $\mu(y)$. Similar to Instance Normalization, these statistics are computed across spatial locations.

Recent studies \cite{zhou2021domain}, have shown that mixing instance-level feature statistics from multiple source domains probabilistically can significantly improve domain generalization. 
This enhancement is achieved by integrating a variety of styles during the training phase, which leads to the development of a model that is more robust and adaptable across different domains. A critical aspect to note is that this mixing of styles and domains occurs during the initial training phase of the model, where the model is trained from scratch with inputs from various domains. This is a crucial distinction, especially in the context of Heterogeneous Face Recognition (HFR), where there is not enough target domain data to do mixing in the training phase. Hence we often start with a face recognition model that is already pre-trained on the source domain.  In the HFR context, we modify this approach by conditionally modulating the feature maps instead of mixing them, and adapting the concept to suit the specific needs of HFR. This conditional modulation in HFR allows the feature statistics of the target modality to adapt, enhancing the model's ability to recognize faces across varied domains without the need for retraining from scratch.

\subsubsection{\fullcaim}

The Adaptive Instance Normalization (AdaIN) module, as previously mentioned, is adept at producing images that match or emulate the style of another image. It is particularly useful in creating images that adopt the style characteristics of a reference image. In the context of Heterogeneous Face Recognition (HFR), the objective is to adjust the style of target modality images so that they align with the style of visible spectrum images. This alignment is crucial to ensure that the final image embeddings are consistent across different modalities. This is of particular importance considering that the pre-trained face recognition network is initially trained on a large dataset of visible spectrum images, making it essential to align the styles between different modalities for effective cross-modal recognition. 

Consider an intermediate feature map in the face recognition network, denoted by $F \in \mathbb{R}^{C \times H \times W}$. Here, $C$, $H$, and $W$ represent the number of channels, height, and width of the feature map, respectively.

For the target modality, we would like to modulate these feature maps such that the output embedding from the network aligns for the source and target modalities. 

To accomplish this, we modulate the intermediate feature map using the \caim block.

\begin{equation}
  \hat{\mathbf{F}} = \mathrm{\caim}(\mathbf{F})
\end{equation}

The \caim block's main component is similar to adaptive instance normalization (AdaIN) particularly in its ability to normalize and modify the style of target images. However, unlike AdaIN which relies on an external style input, our approach derives modulation factors directly from the raw input feature maps using a CNN module. Furthermore, we combine this step in a residual fashion while injecting the \caim block into a pretrained network.

To elaborate further, we first estimate a shared representation from the input feature map by utilizing a shallow CNN network with global average pooling.

\begin{equation}
  \mathbf{\xi_{f}} = \mathrm{GAP}\big(\mathrm{CNN}(\mathbf{F})\big)
\end{equation}

The $\sigma_{f}$ and $\mu_{f}$ parameters are estimated from this shared representation with two fully connected (FC) layers:
\begin{align}
  \mathbf{\sigma_{f}} &= \mathrm{FC}_{\mathrm{\sigma}}(\mathbf{\xi_{f}}) \\
  \mathbf{\mu_{f}} &= \mathrm{FC}_{\mathrm{\mu}}(\mathbf{\xi_{f}})
\end{align}

The estimated parameters are utilized to scale and shift the normalized feature maps:

\begin{equation}
  \mathrm{AIM}(\mathbf{F}) = \mathbf{\sigma_{f}} \left(\frac{\mathbf{F} - \boldsymbol{\mu}(\mathbf{F})}{\boldsymbol{\sigma}(\mathbf{F})}\right) + \mathbf{\mu_{f}}
\end{equation}

To ensure stable training, we incorporate a residual connection in the proposed framework. Additionally, when incorporating this module, a gate is added to activate the module exclusively for the target modality, leaving the feature maps of the source modality unaltered.

The \caim block can be represented as follows:

\begin{equation}
  \mathrm{\caim}(\mathbf{F}, \mathbf{g}) = \mathbf{g} \cdot \mathrm{AIM}(\mathbf{F}) + \mathbf{F}
\end{equation}

Where, $g$ denotes the gate, $\mathbf{g} = 1$ for the target modality, and $\mathbf{g} = 0$ for the source modality (visible images).

\subsection{Face Recognition backbone}

To ensure reproducibility, we used the publicly available pre-trained \textit{Iresnet100} face recognition model provided by Insightface \cite{insightface}. The model was trained on the MS-Celeb-1M dataset \footnote{\url{http://trillionpairs.deepglint.com/data}}, which includes over 70,000 identities. The pre-trained face recognition model accepts three-channel images at a resolution of $112 \times 112$ pixels. Before passing through the FR network, faces are aligned and cropped to ensure eye center coordinates align with predetermined points. In cases where the input is a single-channel image (like NIR or thermal images), the single channel is duplicated across all three channels, without altering the network's architecture.

\subsection{Implementation details}

The \fullcaim (\caim) block employs a contrastive learning approach,  within a Siamese network framework \cite{hadsell2006dimensionality}.  For all experiments, we set the margin parameter to 2.0. The training used the Adam Optimizer with a learning rate of $0.0001$, over 50 epochs, and a batch size of 90. We developed the framework in PyTorch and using the Bob library \cite{bob2017,bob2012} \footnote{\url{https://www.idiap.ch/software/bob/}}. In this setup, the frozen layers of the pre-trained face recognition network are shared between the source and target modalities. The \caim block, inserted between these frozen layers, is operational exclusively for the target modality, activated when the global gate signal is one ($gate=1$). Conversely, for reference channel images (visible spectrum) with $gate=0$, the \caim block essentially acts as a bypass through the residual branch. Only the \caim block's parameters are updated during training. The experiments are reproducible, and the source code and protocols will be made available publicly.

\section{Experiments}
\label{sec:experiments}

This section outlines the outcomes of a comprehensive series of experiments carried out using the \caim framework. Our main objective was to assess the effectiveness of the \caim method in VIS-Thermal \hfr, across various challenging datasets. Furthermore, we compared the performance of the \caim approach against other heterogeneous settings such as VIS-Sketch, VIS-NIR, and VIS-Low Resolution VIS. In all our experiments, we used the standard cosine distance for comparison.

\subsection{Databases and Protocols}

\begin{figure}[t!]
  \centering
  \includegraphics[width=0.99\linewidth]{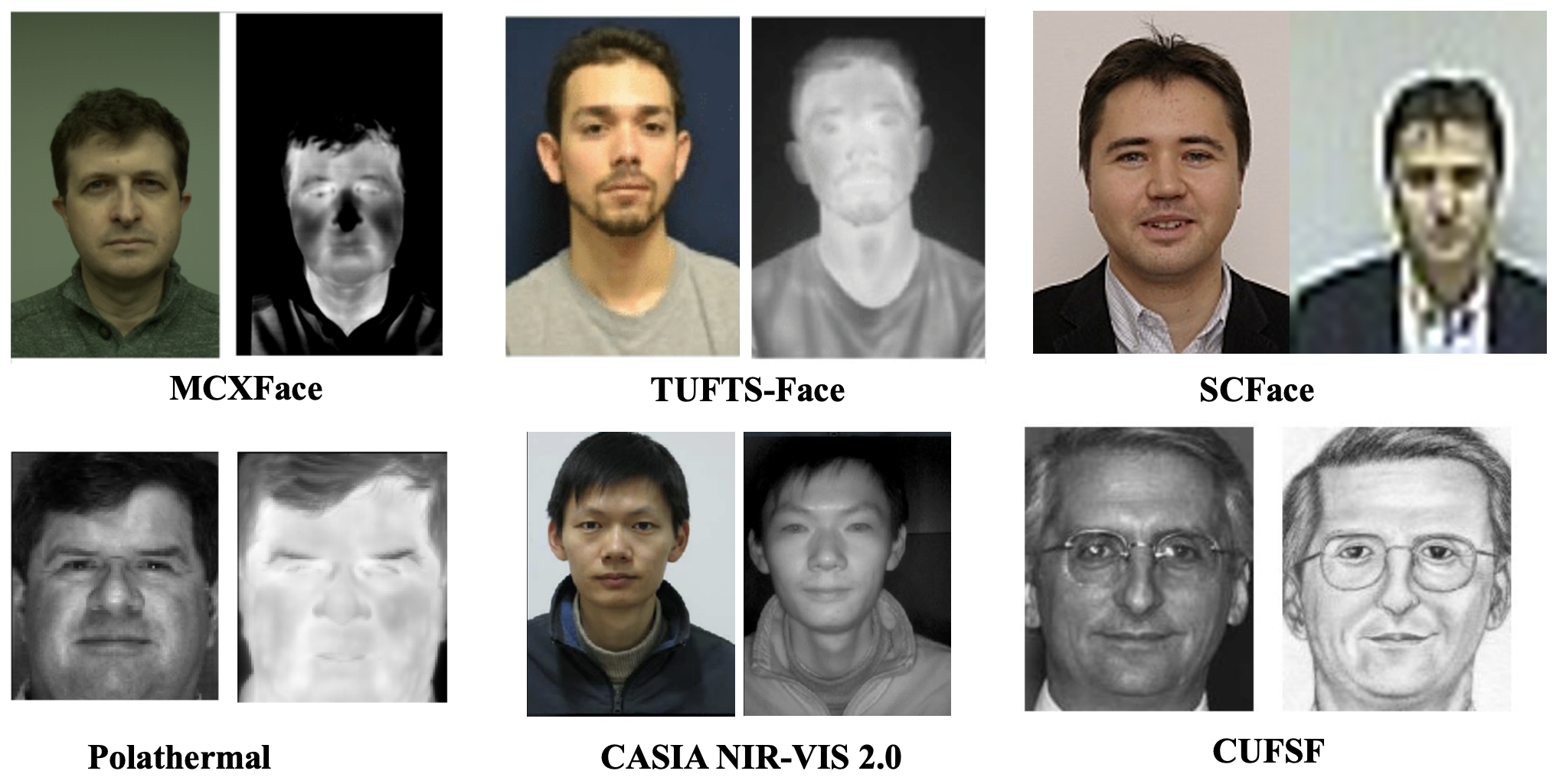}
  \caption{ Sample images from source and target modalities from six different HFR datasets. Images are from MCXFace \cite{george2022prepended}, 
  Tufts Face \cite{panetta2018comprehensive},
  SCFace \cite{grgic2011scface},
  Polathermal  \cite{hu2016polarimetric}
  CASIA NIR-VIS 2.0 \cite{li2013casia},
  CUHK Face Sketch FERET Database (CUFSF) \cite{zhang2011coupled} respectively.
  }
  \label{fig:datasets}
\end{figure}

The following section describes the datasets used (Fig. \ref{fig:datasets}) in the evaluations.

\textbf{Tufts face dataset:} The Tufts Face Database \cite{panetta2018comprehensive} is a collection of face images from various modalities for the \hfr task. For our evaluation of VIS-Thermal \hfr performance, we use the thermal images provided in the dataset. The dataset comprises 113 identities, consisting of 39 males and 74 females from different demographic regions, and includes images from different modalities for each subject. We adopt the same procedure as in \cite{fu2021dvg}, selecting 50 identities at random for the training set and using the remaining subjects for the test set. We report Rank-1 accuracies and Verification rates at false acceptance rates (FAR) of 1\% and 0.1\% for comparison.

\textbf{MCXFace Dataset:} The MCXFace Dataset \cite{george2022prepended} includes images of 51 individuals captured in various illumination conditions and three distinct sessions using different channels. The channels available include RGB color, thermal, near-infrared (850 nm), short-wave infrared (1300 nm), Depth, Stereo depth, and depth estimated from RGB images. All channels are spatially and temporally registered across all modalities. Five different folds were created for each of the protocols by randomly dividing the subjects into $train$ and $dev$ partitions. Annotations for the left and right eye centers for all images are also included in the dataset. We have performed the evaluations on the challenging ``VIS-Thermal'' protocols of this dataset.

\textbf{Polathermal dataset:} The Polathermal dataset \cite{hu2016polarimetric} is an \hfr dataset collected by the U.S. Army Research Laboratory (ARL). It contains polarimetric LWIR imagery together with color images for 60 subjects. The dataset has conventional thermal images and polarimetric images for each subject. For our experiments, we use conventional thermal images and follow the five-fold partitions introduced in \cite{de2018heterogeneous}. Specifically, 25 identities are used for training, while the remaining 35 identities are used for testing. We report the average Rank-1 identification rate from the evaluation set of the five folds.

\textbf{SCFace dataset:} The SCFace dataset \cite{grgic2011scface} consists of high-quality enrollment images for face recognition, while the probe samples are low-quality images from various surveillance scenarios captured by different cameras. There are four different protocols in the dataset, based on the quality and distance of the probe samples: close, medium, combined, and far, with the ``far'' protocol being the most challenging. In total, the dataset contains 4,160 static images from 130 subjects (captured in both visible and infrared spectra).

\textbf{CUFSF dataset:} The CUHK Face Sketch FERET Database (CUFSF) \cite{zhang2011coupled} consists of 1194 faces from the FERET dataset \cite{phillips1998feret}, where each face image has a corresponding sketch drawn by an artist. Due to the exaggerations in the sketches, this dataset poses a challenge for the \hfr task. Following \cite{fang2020identity}, we use 250 identities for training the model and reserve the remaining 944 identities for testing. The Rank-1 accuracies are reported for comparison.

\textbf{CASIA NIR-VIS 2.0 dataset:} The CASIA NIR-VIS 2.0 Face Database \cite{li2013casia}, contains images taken under both the visible spectrum and near-infrared lighting conditions, with 725 distinct individuals. For every person in the dataset, there are 1-22 visible spectrum photos and 5-50 near-infrared (NIR) photos. The given experimental protocols utilize a 10-fold cross-validation method, wherein 360 identities are set aside for training. The evaluation's gallery and probe set comprise 358 distinct individuals. The training and testing sets have entirely separate identities. Experiments are carried out in each fold and the mean and standard deviation of the performance metrics are reported.

\subsection{Metrics}

We evaluate the models using various performance metrics that are commonly used in previous literature, including Area Under the Curve (AUC), Equal Error Rate (EER), Rank-1 identification rate, and Verification Rate at different false acceptance rates (0.01\%, 0.1\%, 1\%, and 5\%).

\subsection{Experimental results}

The experiments performed in the different datasets and the results are discussed in this section. For comparison, we compared the results of \caim against the paper baselines reported in \cite{george2022prepended}.

\subsubsection{\textbf{Experiments with Tufts face dataset}}

The performance of the \caim method and other state-of-the-art techniques in the VIS-Thermal protocol of the Tufts face dataset is presented in Table \ref{tab:tufts}. This dataset is very challenging due to variations in pose and other factors. The extreme yaw angles present in the dataset cause a decline in the performance of even visible spectrum face recognition systems, along with a similar decline in \hfr performance. Despite this challenge, the \caim approach achieves the best verification rate and ranks second in Rank-1 accuracy (73.07\%), following DVG-Face \cite{fu2021dvg}. These results demonstrate the effectiveness of the proposed method.

\begin{table}[h]
  \centering
  \caption{Experimental results on VIS-Thermal protocol of the Tufts Face dataset.}
  \label{tab:tufts}
  \resizebox{0.95\columnwidth}{!}{
  \begin{tabular}{lccc}
    \toprule
    \textbf{Method} & \textbf{Rank-1} & \textbf{VR@FAR=1$\%$} & \textbf{VR@FAR=0.1$\%$}  \\ \midrule
      LightCNN \cite{Wu2018ALC} & 29.4 & 23.0 & 5.3 \\
      DVG \cite{fu2019dual} & 56.1 & 44.3 & 17.1 \\
      DVG-Face \cite{fu2021dvg} & \textbf{75.7} & 68.5 & 36.5 \\ 
      DSU-Iresnet100 \cite{george2022prepended} & 49.7 & 49.8 & 28.3 \\   
      
      PDT \cite{george2022prepended} & 65.71 & 69.4 & 45.5 \\
      MAMCO-HFR \cite{liu2023modality} & - & 68.8 &- \\ \midrule
      \rowcolor{Gray}
      \textbf{\caim (Proposed)} & 73.07 &\textbf{76.81} & \textbf{46.94} \\
      \bottomrule
  
  \end{tabular}
  }
\end{table}

\subsubsection{\textbf{Experiments with MCXFace dataset}}

Table \ref{tab:mcxface} presents the average performance across five folds for the VIS-Thermal protocols in the MCXFace dataset. The reported values are the mean of the five folds in the dataset. The baseline model shown corresponds to the performance of the pretrained \textit{Iresnet100} FR model directly on the thermal images.  It can be seen that the proposed \caim approach achieves the best performance compared to other methods with an average Rank-1 accuracy of 87.24 \%.

\begin{table}[h]
  \caption{Performance of the proposed approach in the VIS-Thermal protocol of  MCXFace dataset, the Baseline is a pre-trained \textit{Iresnet100} model. }
  \label{tab:mcxface}
  \centering
  \resizebox{0.98\columnwidth}{!}{%
  \begin{tabular}{lrrr}
  \toprule
  \textbf{Method} & \textbf{AUC}   & \textbf{EER}   & \textbf{Rank-1}   \\ \midrule
 Baseline & 84.45 $\pm$ 3.70  & 22.07 $\pm$ 2.81 & 47.23 $\pm$ 3.93    \\
DSU-Iresnet100 \cite{george2022prepended} & 98.12 $\pm$ 0.75 & 6.58 $\pm$ 1.35 & 83.43 $\pm$ 5.47 \\

PDT \cite{george2022prepended}      & 98.43 $\pm$ 0.78  &  6.52 $\pm$ 1.45  & 84.52 $\pm$ 5.36   \\ \midrule

\rowcolor{Gray}
\textbf{\caim (Proposed)}  & \textbf{98.97 $\pm$ 0.24} & \textbf{5.05 $\pm$ 0.91} & \textbf{87.24$\pm$2.75} \\
\bottomrule
  \end{tabular}
  }
  \end{table}

\subsubsection{\textbf{Experiments with Polathermal dataset}}
We have performed experiments in the thermal to visible recognition scenarios in the Polathermal dataset and the results are presented in Table \ref{tab:polathermal}. The table shows the average Rank-1 identification rate in the five protocols of the Polathermal `thermal to visible protocols' (using the reproducible protocols in \cite{de2018heterogeneous}). The proposed \caim approach achieves an average Rank-1 accuracy of 95.00\% with a standard deviation of (1.63\%), only second to the PDT approach \cite{george2022prepended}. 

\begin{table}[ht]
\caption{Pola Thermal - Average Rank-1 recognition rate}
\label{tab:polathermal}
\begin{center}
  \resizebox{0.7\columnwidth}{!}{
  \begin{tabular}{lr}
    \toprule
    \textbf{Method} & \textbf{Mean (Std. Dev.)} \\ \midrule
    
    DPM in \cite{hu2016polarimetric}   & 75.31 \% (-) \\ 
    CpNN in \cite{hu2016polarimetric}  & 78.72 \% (-) \\ 
    PLS in \cite{hu2016polarimetric}   & 53.05\% (-)  \\  \midrule

    LBPs + DoG in \cite{liao2009heterogeneous} & 36.8\% (3.5) \\ 
    ISV in \cite{de2016heterogeneous}       & 23.5\% (1.1) \\ 
    GFK in \cite{sequeira2017cross}             & 34.1\% (2.9) \\ 

    DSU(Best Result) \cite{de2018heterogeneous} & 76.3\% (2.1) \\

    DSU-Iresnet100 \cite{george2022prepended} & 88.2\% (5.8) \\
    PDT \cite{george2022prepended} & \textbf{97.1\% (1.3)} \\ \midrule
    \rowcolor{Gray}
    \textbf{\caim (Proposed)} & 95.00\% (1.63) \\
    \bottomrule 
  \end{tabular}
  }
\end{center}
\end{table}

\subsubsection{\textbf{Experiments with SCFace dataset}}

We conducted a series of experiments on the SCFace dataset to evaluate the performance of the proposed approach using the visible images protocol. The dataset presents a heterogeneity challenge due to the quality disparity between the gallery (high-resolution mugshots) and probe (low-resolution surveillance camera) images. The results are presented in Table \ref{tab:scface} and are based on the evaluation set of the standard protocols. The baseline model employed in this experiment is a pre-trained \textit{Iresnet100} model, while the proposed \caim model is trained using contrastive training. It can be seen that the performance of the baseline model improves with the proposed approach in most of the cases. In particular, the improvement is more significant in the ``far'' protocol where the quality of the probe images is very low. The \caim module helps in adapting the intermediate feature map so that the \hfr framework is invariant to quality and resolution, leading to improved results compared to the baseline. The proposed method achieves comparable performance to the PDT approach in this dataset.

\begin{table}[h]
  \caption{Performance of the proposed approach in the SCFace dataset, the Baseline is a pretrained \textit{Iresnet100} model. }
  \label{tab:scface}
  \centering
  \resizebox{0.98\columnwidth}{!}{%
  \begin{tabular}{lcrrrr}
  \toprule
  \textbf{Protocol}             & \textbf{Method} & \textbf{AUC}   & \textbf{EER}   & \textbf{Rank-1}    & \begin{tabular}[c]{@{}c@{}} \textbf{VR@}\\\textbf{FAR=0.1\%} \end{tabular} \\ \midrule
  \multirow{3}{*}{Close} & Baseline & 100.0   & 0.00    & \textbf{100.0}   & 100.0                \\
                        &DSU-Iresnet100 \cite{george2022prepended} & 100.0 & 0.00 & \textbf{100.0} & 100.0    \\

                             &  PDT \cite{george2022prepended}  & 100.0   &  0.00    &  \textbf{100.0}   & 100.0                    \\ 
                             & \cellcolor{Gray} \textbf{\caim (Proposed)} &  \cellcolor{Gray} 100.0 &  \cellcolor{Gray} 0.01 &   \cellcolor{Gray} \textbf{100.0} &  \cellcolor{Gray} 100.0 \\ \midrule

\multirow{3}{*}{Medium}   & Baseline & 99.81 & 2.33 & 98.60  & 92.09              \\
&DSU-Iresnet100 \cite{george2022prepended} & 99.95 & 1.39 & 98.98 & 93.25 \\

                             & PDT \cite{george2022prepended} & 99.96 &  0.93 &  \textbf{99.07} & 95.81                 \\ 
                             & \cellcolor{Gray} \textbf{\caim (Proposed)} &  \cellcolor{Gray} 99.92 &  \cellcolor{Gray} 1.86 &   \cellcolor{Gray} 98.60 &  \cellcolor{Gray} 94.88 \\ \midrule
\multirow{3}{*}{Combined}    & Baseline & 98.59 & 6.67 & 91.01 & 77.67          \\
                                &DSU-Iresnet100 \cite{george2022prepended} & 98.91 & 4.96 &92.71 & 80.93  \\
                             & PDT \cite{george2022prepended} & 99.06 & 4.50  & 93.18 & 82.02           \\  
                             & \cellcolor{Gray} \textbf{\caim (Proposed)} &  \cellcolor{Gray} 99.58 &  \cellcolor{Gray} 3.24 &  \cellcolor{Gray} \textbf{94.57} &  \cellcolor{Gray} 84.65 \\    \midrule
\multirow{3}{*}{Far}     & Baseline & 96.59 & 9.37 & 74.42 & 49.77           \\
                                &DSU-Iresnet100 \cite{george2022prepended} & 97.18 & 8.37 & 79.53 & 58.26  \\

                             & PDT \cite{george2022prepended}   &  98.31 & 6.98 &  84.19 & 60.00            \\ 
                             & \cellcolor{Gray} \textbf{\caim (Proposed)} &  \cellcolor{Gray} 98.81 &  \cellcolor{Gray} 5.09 &  \cellcolor{Gray} \textbf{86.05} &  \cellcolor{Gray} 61.86 \\  

  \bottomrule
  \end{tabular}
  }
  \end{table}

\subsubsection{\textbf{Experiments with CUFSF dataset}}

In this section, we present experiments on the challenging task of sketch to photo recognition. We report the Rank-1 accuracies obtained with the baseline and other methods in Table \ref{tab:cufsf} using the protocols outlined in \cite{fang2020identity}. The proposed approach achieves a Rank-1 accuracy of 76.38\%, which is the best among the compared methods. However, the absolute accuracy in sketch to photo recognition is low compared to other modalities. The CUFSF dataset contains viewed hand-drawn sketch images \cite{klum2014facesketchid} that appear holistically similar to the original subjects for humans. Unlike other imaging modalities such as thermal, near-infrared, and SWIR, sketch images may not preserve the discriminative information that a face recognition network seeks, as they contain exaggerations depending on the artist, making them more challenging for \hfr.  Nevertheless, the proposed \caim approach improves the performance significantly.

\begin{table}[ht]
\caption{CUFSF: Rank-1 recognition in sketch to photo recognition}
\label{tab:cufsf}

\begin{center}
\resizebox{0.5\columnwidth}{!}{%
  \begin{tabular}{lrr}
    \toprule
    \textbf{Method} & \textbf{Rank-1} \\ \midrule
    Baseline & 56.57 \\
    IACycleGAN \cite{fang2020identity} &64.94 \\
    DSU-Iresnet100 \cite{george2022prepended} & 67.06 \\ 
    PDT \cite{george2022prepended}  & 71.08 \\ \midrule
    \rowcolor{Gray} 
    \textbf{\caim (Proposed)} & \textbf{76.38} \\
    \bottomrule 
  \end{tabular}

  }
\end{center}
\end{table}

\subsubsection{\textbf{Experiments with CASIA-VIS-NIR 2.0 dataset}}

We conducted experiments using the CASIA-VIS-NIR 2.0 dataset to demonstrate our proposed method's efficiency in various heterogeneous situations, particularly in VIS-NIR recognition. Observing the baselines, there's a smaller domain gap in this case, with some pre-trained FR models trained in the VIS modality achieving reasonable results. Given this, we employ stricter evaluation thresholds, using VR@FAR=0.1$\%$ and VR@FAR=0.01$\%$ for comparisons. The dataset contains 10 sub-protocols, and we report the average and standard deviation across these ten folds. The findings, shown in Tab. \ref{tab:casia}, reveal that our proposed strategy outperforms other state-of-the-art methods. These results showcase the adaptability of our framework across diverse heterogeneous scenarios.

\begin{table}[h]
  \centering
  \caption{Experimental results on CASIA NIR-VIS 2.0.}
  \label{tab:casia}
  \resizebox{0.47\textwidth}{!}{
  \begin{tabular}{l|ccc}
    \toprule
      Method & Rank-1 & VR@FAR=0.1$\%$ & VR@FAR=0.01$\%$ \\
      \midrule
      IDNet \cite{Reale2016SeeingTF} & 87.1$\pm$0.9 & 74.5 & - \\
      HFR-CNN \cite{saxena2016heterogeneous} & 85.9$\pm$0.9 & 78.0 & - \\
      Hallucination \cite{Lezama2017NotAO} & 89.6$\pm$0.9 & - & - \\
      TRIVET \cite{XXLiu:2016} & 95.7$\pm$0.5 & 91.0$\pm$1.3 & 74.5$\pm$0.7 \\
      W-CNN \cite{DBLP:journals/corr/abs-1708-02412} & 98.7$\pm$0.3 & 98.4$\pm$0.4 & 94.3$\pm$0.4 \\
      PACH \cite{duan2019pose} & 98.9$\pm$0.2 & 98.3$\pm$0.2 & - \\
      RCN \cite{deng2019residual} & 99.3$\pm$0.2 & 98.7$\pm$0.2 & - \\
      MC-CNN \cite{8624555} & 99.4$\pm$0.1 & 99.3$\pm$0.1 & - \\
      DVR \cite{XWu:2019}  & 99.7$\pm$0.1 & 99.6$\pm$0.3 & 98.6$\pm$0.3 \\
      
      DVG \cite{fu2019dual} & 99.8$\pm$0.1 & 99.8$\pm$0.1 & 98.8$\pm$0.2 \\
      DVG-Face \cite{fu2021dvg} & 99.9$\pm$0.1 & 99.9$\pm$0.0 & 99.2$\pm$0.1 \\       
      PDT \cite{george2022prepended} & 99.95$\pm$0.04 & 99.94$\pm$0.03 & 99.77$\pm$0.09 \\
      MAMCO-HFR\cite{liu2023modality} & 99.9$\pm$0.1 & 99.8$\pm$0.1 & - \\ \hline
\rowcolor{Gray}
      \textbf{CAIM (Proposed)} & 99.96$\pm$0.02 &99.95$\pm$0.02 & 99.79$\pm$0.11 \\ \bottomrule 
  \end{tabular}}
\end{table}

\begin{table}[!htb]
  \caption{Performance with different number of \caim blocks. 1-5 indicates the \caim module is inserted in all blocks from first to fifth layers. Experiment performed in Tufts face dataset. }
  \centering
  \resizebox{0.99\columnwidth}{!}{%
\begin{tabular}{lrrrrr}
  \toprule
  \textbf{Layers} &             \textbf{AUC} &             \textbf{EER} &              \textbf{Rank-1} &          \textbf{VR(0.1\% FAR)} &            \textbf{VR(1\% FAR)}    \\
  \midrule
  1 &  91.28 &  17.10 &  49.19 &    3.34 &  49.17  \\
  1-2 &  94.91 &  11.35 &  64.45 &   39.70 &  67.72  \\
  \rowcolor{Gray}
  1-3 &  \textbf{97.01} &   \textbf{8.53} &  \textbf{73.07} &   \textbf{46.94} &  \textbf{76.81} \\
  1-4 &  96.18 &   9.28 &  68.76 &   45.08 &  72.36 \\
  1-5 &  95.73 &  10.76 &  69.30 &   33.40 &  71.61 \\
  \bottomrule
  \end{tabular}
  }
  \label{tab:ablation_layer}
\end{table}

\subsection{Ablation Studies}

In this subsection, we conduct a series of ablation studies to better understand the efficacy of various components and to assess the generalizability of the CAIM approach.

\subsubsection{Effect of number of CAIM blocks}

To understand the effect of having a different number of \caim blocks, we performed a set of experiments in the Tufts face dataset by inserting a different number of \caim blocks in the pre-trained FR network.  We start by placing only one \caim block after the first block of the pre-trained FR layer. Then we increased the number of \caim blocks from one to 5.  
The results of this experiment are presented in Table \ref{tab:ablation_layer}. Our analysis reveals that adapting lower layers is effective in minimizing the domain gap, as the high-level facial structure is consistent across various modalities. In this case,  adding three \caim blocks achieved the best performance (this setting is used in all other experiments). Conversely, adapting more layers does not bring significant improvements as they are more task-specific. In our case, the task is face recognition which is the same for both source and target modalities.

The optimal number of layers to adapt can vary depending on the specific modality and architecture, but we have observed that adapting layers ``1-3'' generally yields satisfactory results across a diverse range of modalities. Consequently, we have consistently applied these settings in all our experiments, though they may not be optimal. Conducting additional experiments to determine the optimal number of layers for each dataset and architecture could potentially enhance performance. To evaluate this, we have evaluated our approach on the CUFSF dataset with a different number of layers, and the results are shown in Table \ref{tab:ablation_layer_iresnet100_cufsf}. The results indicate a modest improvement in rank-1 accuracy when layers ``1-4'' are adapted. However, modifying additional layers poses risks of overfitting and increased computational overhead (Table \ref{tab:computational_complexity}). 

\begin{table}[!htb]
  \caption{Performance with different number of CAIM blocks. 1-5 indicates the CAIM module is inserted in all blocks from the first to fifth layers. The experiment was performed in CUFSF face dataset for iresnet100 model.} 
  \centering
  \resizebox{0.99\columnwidth}{!}{%
\begin{tabular}{lrrrrr}
  \toprule
  \textbf{Layers} &             \textbf{AUC} &             \textbf{EER} &              \textbf{Rank-1} &          \textbf{VR(0.1\% FAR)} &            \textbf{VR(1\% FAR)}    \\
  \midrule

1   &  99.22 &  4.13 &  65.89 &   69.49 &  89.30 \\
1-2 &  99.53 &  3.07 &  69.28 &   72.78 &  90.57 \\
1-3 &  99.78 &  2.01 &  76.38 &   81.25 &  95.55 \\
1-4 &  99.72 &  2.33 &  76.69 &   80.72 &  95.55 \\
1-5 &  99.72 &  2.63 &  76.17 &   79.34 &  94.28 \\
\bottomrule
\end{tabular}
  }
  \label{tab:ablation_layer_iresnet100_cufsf}
\end{table}

Additionally, we conducted experiments using the ElasticFace model on the Tufts face dataset to determine the optimal number of layers. In this scenario, adapting layers ``1-4'' proved more effective than just ``1-3''. This suggests that the number of layers to tune can be further optimized for separate models and datasets. Nonetheless, adapting ``1-3" layers provides a reasonable trade-off in terms of performance and computational overhead. 

\begin{table}[!htb]
  \caption{Performance with different number of CAIM blocks. 1-5 indicates the CAIM module is inserted in all blocks from the first to fifth layers. The experiment was performed in Tufts face dataset for ElasticFace model.} 
  \centering
  \resizebox{0.99\columnwidth}{!}{%
  
\begin{tabular}{lrrrrr}
  \toprule
  \textbf{Layers} &             \textbf{AUC} &             \textbf{EER} &              \textbf{Rank-1} &          \textbf{VR(0.1\% FAR)} &            \textbf{VR(1\% FAR)}    \\
  \midrule

1   &  87.50 &  20.41 &  48.11 &   25.23 &  46.57 \\
1-2 &  93.59 &  13.54 &  61.76 &   31.54 &  59.18 \\
1-3 &  95.24 &  10.39 &  73.43 &   50.65 &  73.65 \\
1-4 &  96.04 &  10.20 &  71.81 &   56.77 &  74.03 \\
1-5 &  94.34 &  13.36 &  59.25 &   29.50 &  60.30 \\
\bottomrule

  \end{tabular}
  }
  \label{tab:ablation_layer_elastic}
\end{table}

\subsubsection{Effectiveness of components of CAIM block}

Further to understand the effectiveness of the conditional operation, we conducted experiments using the \textbf{AIM} and Instance Norm (IN) modules in an unconditional manner. These experiments were conducted using the Tufts-face dataset, with the results shown in Table \ref{tab:tufts_uncond}. The conditional path in \caim keeps the original performance on the source modality intact and prevents catastrophic forgetting when adapted to an extra modality. It can be seen that an unconditional integration of the block violates this premise and leads to inferior performance. These results underline both the effectiveness and necessity of the conditional operation.

\begin{table}[h]
  \centering
  \caption{Ablation experiments on Tufts Face dataset with unconditional block.}
  \label{tab:tufts_uncond}
  \resizebox{0.95\columnwidth}{!}{
  \begin{tabular}{lccc}
    \toprule
    \textbf{Method} & \textbf{Rank-1} & \textbf{VR@FAR=1$\%$} & \textbf{VR@FAR=0.1$\%$}  \\ \midrule

AIM & 6.82  &  3.71 &   0.19 \\

IN &36.27  &  17.44 &   3.53 \\ \midrule
\rowcolor{Gray}
      \textbf{\caim (Proposed)} & 73.07 &\textbf{76.81} & \textbf{46.94} \\
      \bottomrule
  
  \end{tabular}
  }
\end{table}

\subsubsection{Experiment with another FR model}

To evaluate the effectiveness of the approach with models trained with different loss functions, we have performed experiments with models trained with ArcFace \cite{deng2019arcface} and ElasticFace \cite{boutros2022elasticface} loss functions. We use the same \textit{iresnet100} architecture for both of these models. The performance of the models is shown in Table \ref{tab:tufts_fr_models}. It can be seen that both models perform comparably, showcasing the effectiveness of the approach. Also, it is to be noted that, despite the original models being trained using different loss functions, the learning phase of the \caim module is the same as described in the previous section.  
\begin{table}[h]
  \centering
  \caption{Ablation experiments on Tufts Face dataset with ArcFace and ElasticFace.}
  \label{tab:tufts_fr_models}
  \resizebox{0.95\columnwidth}{!}{
  \begin{tabular}{lccc}
    \toprule
    \textbf{Method} & \textbf{Rank-1} & \textbf{VR@FAR=1$\%$} & \textbf{VR@FAR=0.1$\%$}  \\ \midrule

     ElasticFace \cite{boutros2022elasticface} + CAIM &\textbf{73.43}  &73.65 & \textbf{50.65}  \\
      ArcFace \cite{deng2019arcface} + CAIM & 73.07 &\textbf{76.81} & 46.94 \\
      \bottomrule
  
  \end{tabular}
  }
\end{table}

\subsubsection{Computational Complexity}

We have evaluated the computational load of the CAIM approach applied to the iresnet100 architecture, particularly measuring the overall computation in terms of floating point operations (GFLOPS) and the number of parameters (expressed in millions of parameters - MPARAMS). The results presented in Table \ref{tab:computational_complexity} show that the additional computational load and parameters required by the CAIM approach are minimal. To be more specific, adapting layers ``1-3'' results in a mere 0.6\% increase in the number of parameters and an 8.6\% rise in computational requirements, while converting the FR model to an HFR one. 

\begin{table}[!htb]
  \caption{Computational complexity of the CAIM approach with different number of layers in terms of floating point operations (GFLOPS) and the number of parameters (expressed in millions of parameters - MPARAMS). }
  \label{tab:computational_complexity}
  \centering
  \resizebox{0.7\columnwidth}{!}{%
\begin{tabular}{lcc}
\toprule
{} &  GFLOPS &  MPARAMS \\
\midrule
iresnet100           &    2.42 &    65.15 \\ \hline
iresnet100 + CAIM(1)   &    2.56 &    65.22 \\
iresnet100 + CAIM(1-2) &    2.59 &    65.30 \\
iresnet100 + CAIM(1-3) &    2.63 &    65.58 \\
iresnet100 + CAIM(1-4) &    2.66 &    66.73 \\
iresnet100 + CAIM(1-5) &    2.69 &    71.32 \\
\bottomrule
\end{tabular}
}
\end{table}

\section{Discussions}
\label{sec:discussions}
We introduced a new strategy that adapts feature maps of the target modality to align with the style of visible images, thereby effectively reducing the domain gap between different image modalities. To achieve this, we introduce a novel module called \caim that can be inserted into a pre-trained FR model, which enables the conversion of a face recognition model to an \hfr model. Our experimental results demonstrate the effectiveness and robustness of our proposed approach, with state-of-the-art performance achieved in various \hfr benchmarks. In five out of six datasets, the proposed approach outperforms all other approaches compared. Our method shows superior adaptability in the feature space compared to PDT \cite{george2022prepended}, whose transformations are constrained by the PDT block’s receptive field, making our framework more flexible. Our approach can convert an FR model to an HFR model with less than 10\% additional compute. The proposed approach can be extended to newer FR architectures, and can also be improved by better training methods.

\section{Conclusions}
\label{sec:conclusions}

In this work, we introduce a novel framework for heterogeneous face recognition by considering different imaging modalities as distinct ``styles''.  Our proposed strategy transforms a conventional face recognition (FR) model into a heterogeneous face recognition (HFR) model by aligning the style of the target modality feature maps with that of visible images. To accomplish this, we introduce a novel network module named ``CAIM'',  which can be seamlessly integrated between the frozen layers of a pre-trained FR network. This new CAIM module is trained for HFR in a contrastive learning setup. Our experimental results showcase our method's state-of-the-art performance across several challenging benchmarks. Our approach is versatile and compatible with face recognition models trained using different loss functions. To encourage further research and extensions of our work, we will make the source codes and protocols available publicly.
\ifCLASSOPTIONcompsoc
  \section*{Acknowledgments}
\else
  \section*{Acknowledgment}
\fi

The authors would like to thank the Swiss Center for Biometrics Research and Testing for supporting the research leading to results published in this paper.

\ifCLASSOPTIONcaptionsoff
  \newpage
\fi



%
\bibliographystyle{IEEEtran}

\bibliography{sn-bibliography}

\begin{IEEEbiography}[{\includegraphics[width=1in,height=1.25in, trim={0cm 0.5cm 0cm 0.3cm},clip,keepaspectratio]{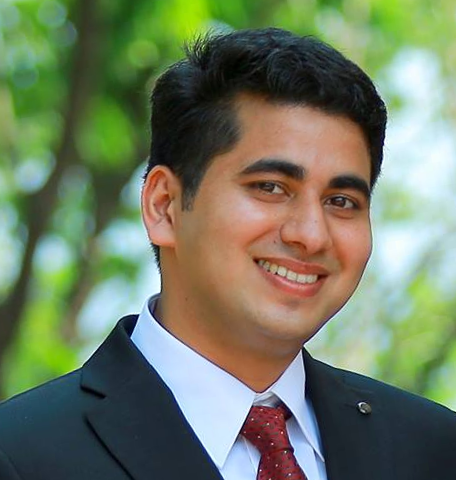}}]{Anjith George} has received his Ph.D. and M-Tech degree from the Department of Electrical Engineering, Indian Institute of Technology (IIT) Kharagpur, India in 2012 and 2018 respectively. After Ph.D, he worked in Samsung Research Institute as a machine learning researcher. Currently, he is a research associate in the biometric security and privacy group at Idiap Research Institute, focusing on developing face recognition and presentation attack detection algorithms. His research interests are real-time signal and image processing, embedded systems, computer vision, machine learning with a special focus on Biometrics.
\end{IEEEbiography}

\begin{IEEEbiography}[{\includegraphics[width=1in,height=1.25in,trim={7cm 0cm 7cm 0.5cm},clip,keepaspectratio]{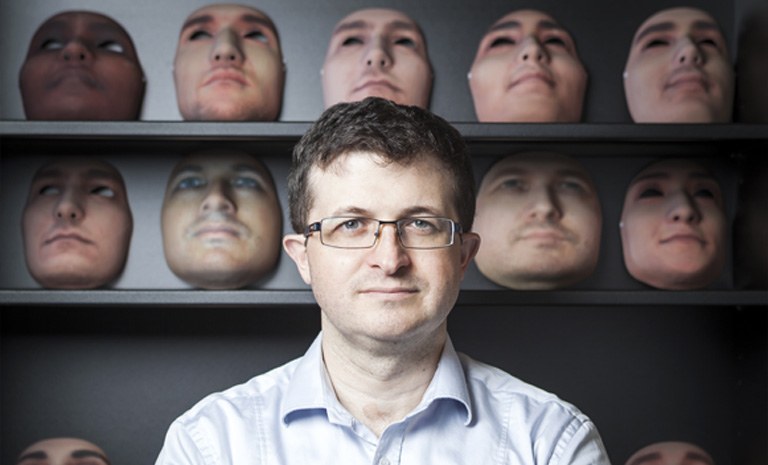}}]{S{\'e}bastien Marcel} heads the Biometrics Security and Privacy group at Idiap Research Institute (Switzerland) and conducts research on face recognition, speaker recognition, vein recognition, attack detection (presentation attacks, morphing attacks, deepfakes) and template protection. He received his Ph.D. degree in signal processing from Universit{\'e} de Rennes I in France (2000) at CNET, the research center of France Telecom (now Orange Labs). He is Professor at the University of Lausanne (School of Criminal Justice) and a lecturer at the  \'{E}cole Polytechnique F{\'e}d{\'e}rale de Lausanne. He is also the Director of the Swiss Center for Biometrics Research and Testing, which conducts certifications of biometric products.
\end{IEEEbiography}


\end{document}